\pdfoutput=1

\documentclass[11pt]{article}

\usepackage[final]{acl}

\usepackage{times}
\usepackage{latexsym}
\usepackage[T1]{fontenc}

\usepackage[utf8]{inputenc}

\usepackage{microtype}

\usepackage{inconsolata}

\usepackage{graphicx}

\usepackage{booktabs} 
\usepackage{multirow}
\usepackage[table]{xcolor}
\newcommand{\gc}{\cellcolor[gray]{0.9}}
\usepackage{tcolorbox}
\usepackage{hyperref}
\usepackage{xspace}
\newcommand{\ie}{{\emph{i.e.}},\xspace}

\newcommand{\eg}{{\emph{e.g.}},\xspace}

\usepackage{amsmath}
\usepackage{amssymb}
\usepackage{mathtools}
\usepackage{amsthm}

\usepackage[capitalize,noabbrev]{cleveref}

\theoremstyle{plain}
\newtheorem{theorem}{Theorem}
\newtheorem{proposition}{Proposition}

\newtheorem{definition}{Definition}

\theoremstyle{definition}
\newtheorem*{remark}{Remark}

\title{Your Language Model Can Secretly Write Like Humans: Contrastive Paraphrase Attacks on LLM-Generated Text Detectors}

\author{
 \textbf{Hao Fang\thanks{Equal Contribution}\textsuperscript{1}},
 \textbf{Jiawei Kong$^*$\textsuperscript{1,2}},
 \textbf{Tianqu Zhuang\textsuperscript{1}},
 \textbf{Yixiang Qiu\textsuperscript{1}},
 \textbf{Kuofeng Gao\textsuperscript{1}}, 
 \\
 \textbf{Bin Chen\thanks{Corresponding Author}\textsuperscript{2,3}},
 \textbf{Shu-Tao Xia\textsuperscript{1,3}},
 \textbf{Yaowei Wang\textsuperscript{2,3}},
 \textbf{Min Zhang\textsuperscript{2}},
\\
 \textsuperscript{1}Tsinghua Shenzhen International Graduate School, Tsinghua University,\\
 \textsuperscript{2}Harbin Institute of Technology, Shenzhen,
 \textsuperscript{3}Pengcheng Laboratory
\\
\small\tt{\{fangh25, kjw25, zhuangtq23, qiu-yx24, gkf21\}@mail.tsinghua.edu.cn, chenbin2021@hit.edu.cn,} 
\\
\small\tt{xiast@sz.tsinghua.edu.cn, wangyw@pcl.ac.cn, zhangmin2021@hitsz.edu.cn}
}

\begin{document}
\maketitle

\begin{abstract}
The misuse of large language models (LLMs), such as academic plagiarism, has driven the development of detectors to identify LLM-generated texts. To bypass these detectors, paraphrase attacks have emerged to purposely rewrite these texts to evade detection. Despite the success, existing methods require substantial data and computational budgets to train a specialized paraphraser, and their attack efficacy greatly reduces when faced with advanced detection algorithms. To address this, we propose \textbf{Co}ntrastive \textbf{P}araphrase \textbf{A}ttack (CoPA), a training-free method that effectively deceives text detectors using off-the-shelf LLMs. The first step is to carefully craft instructions that encourage LLMs to produce more human-like texts. Nonetheless, we observe that the inherent statistical biases of LLMs can still result in some generated texts carrying certain machine-like attributes that can be captured by detectors. To overcome this, CoPA constructs an auxiliary machine-like word distribution as a contrast to the human-like distribution generated by the LLM. By subtracting the machine-like patterns from the human-like distribution during the decoding process, CoPA is able to produce sentences that are less discernible by text detectors. Our theoretical analysis suggests the superiority of the proposed attack. Extensive experiments validate the effectiveness of CoPA in fooling text detectors across various scenarios. The code is available at: \textcolor{magenta}{\url{https://github.com/ffhibnese/CoPA_Contrastive_Paraphrase_Attacks}}
\end{abstract}

\section{Introduction}

Large language models (LLMs), such as GPT-4 and Claude-3.5, have demonstrated remarkable abilities in text comprehension and coherent text generation. These capabilities have driven their widespread applications, including code generation \cite{jiang2024survey} and academic research \cite{chris2022smart}. However, the misuse of LLMs for harmful purposes, such as academic plagiarism and misinformation generation, has raised significant societal concerns regarding safety and ethics~\cite{baofast, wu2025moses, kong2025wolf}. 
In response, various detection methods that leverage the unique characteristics of LLM-generated texts from multiple perspectives have been proposed to mitigate the associated risks.

\begin{figure}[t]
\begin{center}
\includegraphics[width=\linewidth]{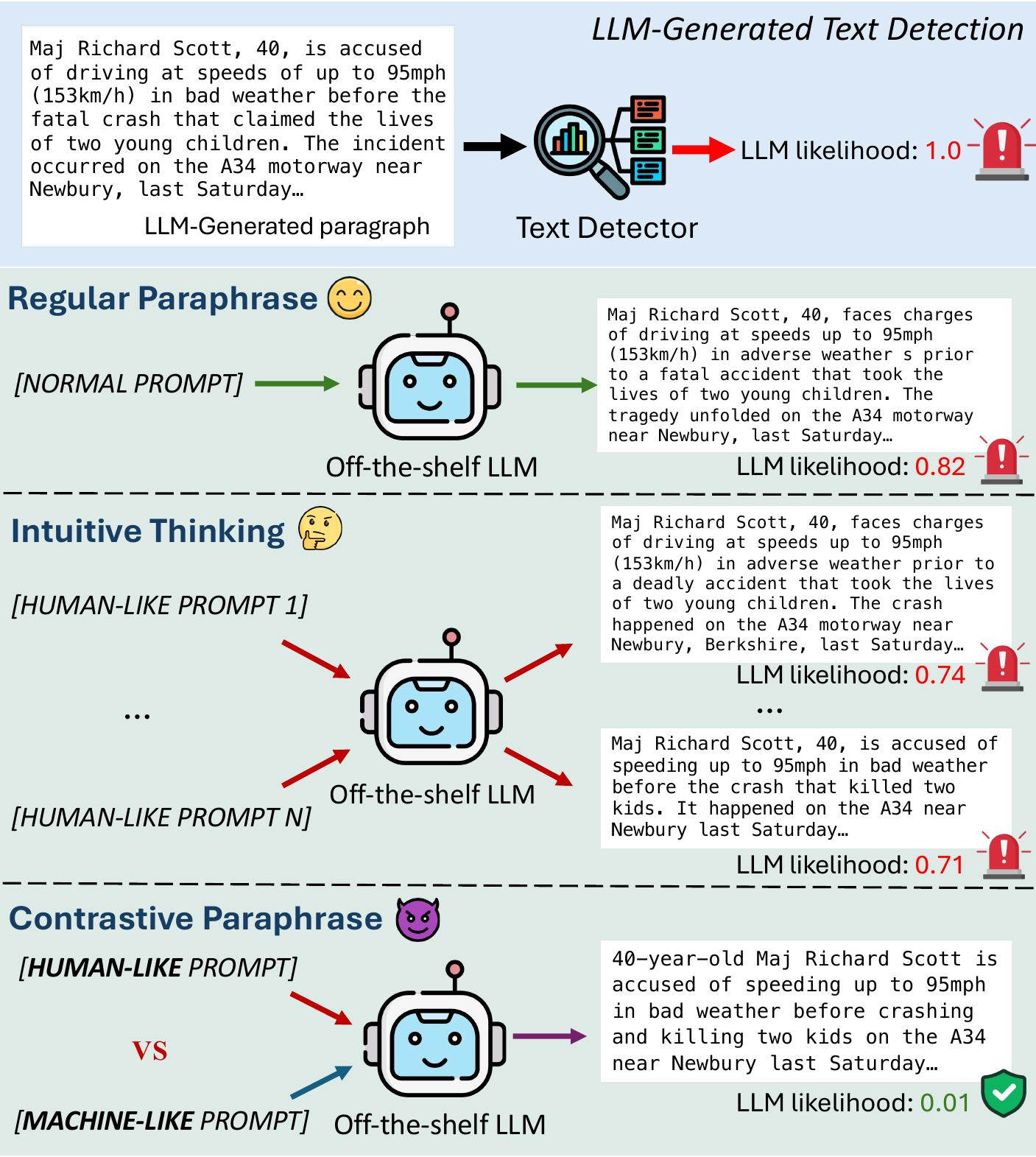}
\end{center}
\vspace{-0.8em}
\caption{Comparison of different paraphrasing strategies. The human-like and machine-like prompts are crafted to guide the LLM in generating human-style and machine-style texts, respectively.}
\label{fig:intro}
\vspace{-0.8em}
\end{figure}

Concurrently, red-teaming countermeasures \cite{krishna2023paraphrasing, shi2024red} have also been introduced to evaluate the reliability of these detection algorithms, which can be broadly categorized into \textit{word-substitution} and \textit{paraphrase} attacks. 
Specifically, word-substitution attacks \cite{shi2024red, wang2024raft} replace specific important words in the fake sentence using candidate words generated by a language model. 
However, they require an additional surrogate model to identify the word importance, and the replacement operation can significantly increase sentence perplexity \cite{wang2024raft}, making the processed texts easily identified by humans. 
In contrast, Dipper \cite{krishna2023paraphrasing} proposes a paraphrase-based attack that rewrites the whole paragraph by modifying the syntax and phrasing to deceive text detectors. This approach does not rely on surrogate models and can preserve sentence perplexity, presenting a more practical and versatile attack strategy. 
Nonetheless, Dipper requires training a large-scale generative language model as the paraphraser, incurring substantial data collection and computational burdens. Moreover, the attack performance is greatly diminished when confronted with more advanced defense strategies such as Fast-DetectGPT \cite{baofast}, as shown Sec. \ref{sec:main_exp}.

In this paper, we build upon the research line of paraphrase attacks and propose a training-free paraphrase approach named Contrastive Paraphrase Attack (CoPA), which aims to elicit human-written word distributions from an off-the-shelf LLM to evade detection. 
Specifically, we revisit the fundamental mechanisms underlying existing detection algorithms and hypothesize that the core principle of effective paraphrase attacks lies in erasing the machine-inherent characteristics within the paragraph while incorporating more human-style features, such as more flexible choices of words and phrases.
Based on this insight, we intuitively seek to craft prompts that alleviate the established statistical constraints of pre-trained LLMs and produce more human-like word distributions, as shown in Fig. \ref{fig:intro}. 
While this strategy exhibits some effectiveness, LLMs pretrained on massive corpora hold a strong tendency to prioritize words with high statistical probability to ensure sentence coherence \cite{baofast, maoraidar}. This inherent bias consistently influences the word choices of generated sentences, irrespective of the input prompts. Therefore, some paraphrased sentences still exhibit certain machine-related characteristics, rendering them highly discernible by detection classifiers.

To address this limitation, we conduct a reverse-thinking analysis. While it is challenging to directly produce highly human-written word distributions that can fully bypass detection, eliciting the opposite machine-like word distributions that contain rich machine-related attributes is considerably easier. These machine-like word probabilities can then serve as negative instances to further purify the obtained human-style distribution for more human-like text generation. 
In light of this consideration, an auxiliary machine-like distribution is constructed as a contrastive reference, which is used to filter out the machine-related concepts from the aforementioned human-like distribution. By sampling from the meticulously adjusted word distribution, CoPA generates more diverse and human-like sentences, which exhibit remarkable effectiveness in deceiving LLM-text detectors.

\textbf{Contributions.} We propose CoPA, a novel paraphrase attack that contrastively modifies the word distribution from an off-the-shelf LLM to rewrite generated texts for enhanced attacks against text detectors.
CoPA eliminates the cumbersome burdens of training a dedicated paraphraser, achieving an efficient and effective attack paradigm. Furthermore, we develop a theoretical framework that substantiates the superiority of CoPA.

To validate the effectiveness, we conduct extensive experiments on 3 long-text datasets with various styles against 8 powerful detection algorithms. 
Compared to baselines, CoPA consistently enhances the attack while maintaining semantic similarity, \eg an average improvement of 57.72\% in fooling rates (at FPR=5\%) for texts generated by GPT-3.5-turbo when against Fast-DetectGPT. 

\section{Related Work}

\subsection{LLM-generated Text Detection}
Existing detection algorithms for AI-generated text can generally be categorized into two types: (i) Training-based detection: These methods typically involve training a binary classification language model. Specifically, OpenAI employs a RoBERTa model \cite{liu2019roberta}  trained on a collection of millions of texts for detection. To enhance the detection robustness, RADAR \cite{hu2023radar} draws inspiration from GANs \cite{goodfellow2020generative} and incorporates adversarial training between a paraphraser and a detector. Additionally, DeTective \cite{guo2024detective} proposes a contrastive learning framework to train the encoder to distinguish various writing styles of texts, and combined with a pre-encoding embedding database for classification. R-detect \cite{songdeep} employs a kernel relative test to judge a text by determining whether its distribution is closer to that of human texts.
Despite the efforts in specific domains, training-based methods are struggling with generalization to unseen language domains, which reduces their practicality and versatility. 
(ii) Zero-shot detection: These methods are training-free and typically focus on extracting inherent features of LLMs' texts to make decisions. GLTR \cite{gehrmann2019gltr} and LogRank\cite{solaiman2019release} leverage the probability or rank of the next token for detection. Since AI-generated text typically exhibits a higher probability than its perturbed version, DetectGPT \cite{mitchell2023detectgpt} proposes the probability curvature to distinguish LLM and human-written text. Building on this, Fast-DetectGPT \cite{baofast} greatly improves the efficiency by introducing conditional probability curvature that substitutes the perturbation step with a more efficient sampling step. TOCSIN \cite{ma2024zero} presents a plug-and-play module that incorporates random token deletion and semantic difference measurement to bolster zero-shot detection capabilities. Other methods explore different characteristics, such as likelihood \cite{hashimoto2019unifying}, N-gram divergence \cite{yangdna}, and the editing distance from the paraphrased version \cite{maoraidar}.

\subsection{Attacks against Text Detectors}
Red-teaming countermeasures have been proposed to stress-test the reliability of detection systems. Early attempts evade detection using in-context learning \cite{lu2023large} or directly fine-tuning the LLM \cite{nicks2023language} under a surrogate detector. Recent advances can be broadly categorized into:  

\textbf{Substitution-based attacks.}
\citet{shi2024red} introduce the substitution-based approach, which minimizes the detection score provided by a surrogate detector by replacing certain words in AI sentences with several synonyms generated by an auxiliary LLM. 
Subsequently, RAFT \cite{wang2024raft} improves the attack performance by introducing an LLM-based scoring model to greedily identify critical words in the machine sentences. 
However, this type of attack relies on an additional surrogate model, and the generated sentences suffer from reduced coherence and fluency, which are easily identifiable by human observations and limit their practical utility in real-world scenarios.  

\textbf{Paraphrasing-based attacks.}
Conversely, Dipper \cite{krishna2023paraphrasing} suggests a surrogate-free approach that can perfectly maintain text perplexity. With a rewritten dataset of paragraphs with altered word and sentence orders, Dipper fine-tunes a T5-XXL  \cite{raffel2020exploring} as a paraphraser to rewrite entire machine paragraphs, which effectively fools text detectors while preserving semantic consistency. 
Based on Dipper, \citet{sadasivan2023can} introduce a recursive strategy that performs multiple iterations of paraphrasing, with slightly degrading text quality while significantly enhancing attack performance. Raidar \cite{maoraidar} proposes a straightforward approach that directly queries an LLM to paraphrase machine text into paragraphs with more human-style characteristics. \citet{shi2024red} suggest a strategy that automatically searches prompts to induce more human-like LLM generations. However, it relies on a strong assumption that a surrogate detector is available. 

This paper follows the more practical and applicable paradigm of surrogate-free paraphrasing attacks and proposes a contrastive paraphrasing strategy, which achieves remarkable efficacy in bypassing LLM-text detection systems.
\section{Method}
In this section, we first present the paradigm of paraphrase attacks. Then, we elaborate on the proposed CoPA that rewrites generated texts using a pre-trained LLM. Finally, we provide a theoretical framework to guarantee the attack effectiveness.

\subsection{Problem Formulation}
We denote the AI-generated text detector weighted by $w$ as $D_w: \mathcal{Y} \rightarrow [0, 1] $, where $\mathcal{Y}$ denotes the text domain. The detector \( D_w \) maps text sequences \( y \in \mathcal{Y} \) to corresponding LLM likelihood scores, where higher values indicate a greater probability of being LLM-generated. For an LLM pre-trained on extensive corpora, the resulting texts exhibit significant writing styles, including word preferences and coherent syntactic structures, which have been leveraged in previous studies to develop various text detectors. 
To evade detection, a malicious paraphrase attacker aims to reduce these machine-related concepts within the machine-generated texts to obtain paraphrased variants that can effectively mislead \( D_w \) into outputting lower LLM likelihood scores. 
Before delving into the proposed method, we first review the token generation paradigm of the LLM inference.
\begin{figure*}[t]
\begin{center}
\includegraphics[width=0.94\linewidth]{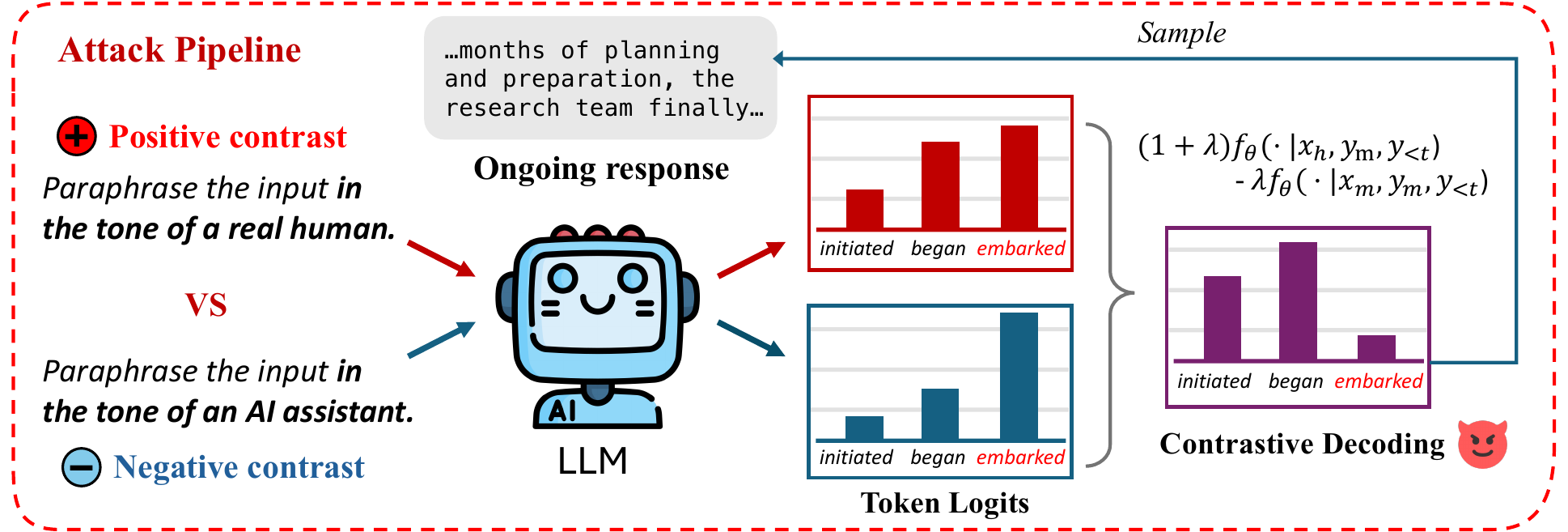}
\end{center}
\vspace{-0.5em}
\caption{Overview of the proposed CoPA. The contrastive paraphrasing successfully penalizes the LLM-preferred word `\textit{embarked}' \cite{liang2024monitoring} and encourages more flexible word choices for next-token sampling.}
\label{fig:pipeline}
\end{figure*}
We consider utilizing an off-the-shelf LLM as the paraphraser. Given an input instruction \( x \), a machine text $y_m$ and a pre-trained LLM $f_\theta(\cdot)$ parameterized by \(\theta\), \( f_\theta \) generates the paraphrased paragraph \( y \) by producing tokens in an autoregressive manner. At each timestep \( t \), $f_\theta(\cdot)$ samples the next token \( y_t \) from the conditional probability distribution:  
\begin{equation}
\begin{split}
y_t\sim p_{\theta}(\cdot| x, y_m, y_{<t}) \propto \exp{f_\theta(\cdot|x, y_m, y_{<t})},
\label{eq:LLM_sample}
\end{split}
\end{equation}
where \( y_{<t} \) denotes the previously generated sequence.  Particularly, the probability of a generated text \( y \) with length \( l \) can be expanded as the multiplication of conditional probabilities:  
\begin{equation}
\begin{split}
q_{\theta}(y) = \prod_{t=1}^l p_{\theta}(y_t |x, y_m, y_{<t}),
\label{eq:LLM_sentence_sample}
\end{split}
\end{equation}
where $q_{\theta}(\cdot)$ denotes the text probability distribution. Based on the chain rule in Eq. (\ref{eq:LLM_sentence_sample}), the sentence-level paraphrasing problem can be further reformulated as a token-level selection task, \ie design algorithms to adequately penalize the probability of machine-favored tokens and inspire more human-like word choices for generating sentences able to confuse the text detector $D_w(\cdot)$. 
In addition to outstanding attack performance, the revised sentences should preserve the original semantics and exhibit a high degree of coherence to ensure text quality.

\subsection{Contrastive Paraphrase Attacks}
Building upon the preceding analysis, an intuitive approach is to devise a prompt \( x_{h} \) that can elicit more diverse token distributions $p_h'$ from LLM $f_{\theta}(\cdot)$ to simulate authentic human-written distribution $p_h$. While this strategy achieves some success, we find that the inherent statistical priors of language models persistently impose constraints on the output distributions. As a result, this leads to unstable outcomes, with some revised sentences still exhibiting sufficient machine-related patterns and remaining highly detectable (see Appendix \ref{examples_human_only}). 

To achieve more effective and stable attacks, we carefully examine this issue and identify the following two critical considerations.
(1) The current strategy essentially operates within the input space (\ie modify input prompts) to indirectly influence the output token distribution, which remains inevitably constrained by the prior knowledge encoded in the LLM. Instead, directly manipulating the output distribution is a potentially more promising alternative. (2) Generating word distributions that can fully deceive detection models is challenging; however, it is much easier to generate word distributions that are highly detectable. From a reverse-thinking perspective, those LLM-favored tokens are also considerably valuable since they encapsulate rich machine-style features and can serve as negative examples for contrastive references.
Based on these insights, we propose our Contrastive Paraphrase Attack, a novel approach that directly employs a dynamic adjustment to the output word distributions during LLM decoding. Figure \ref{fig:pipeline} illustrates the core pipeline of CoPA.
Apart from the prompt $x_{h}$ for human-style distributions $p_h'$, we also construct a comparative machine prompt $x_{m}$ to elicit machine-preferred word choices $p_m$. By contrasting human-like and machine-like token distributions, CoPA refines the probabilities in the decoding process and encourages generations of texts with enhanced human-written resemblance. 
Formally, the contrastively purified token distribution at timestep $t$ can be expressed as:  
\begin{equation}
\begin{split}
p_{c}&(\cdot|x_h, x_m, y_m, y_{<t}) \propto \exp\Big(( 1+\lambda) \\&f_\theta( \cdot|x_h, y_m, y_{<t} ) - \lambda f_\theta(\cdot|x_m, y_m, y_{<t}) \Big), 
\label{eq:contrastive_decode}
\end{split}
\end{equation}
where \( p_{\text{c}} \) represents the contrastive token distributions and \(\lambda\) is the scaling parameter that controls the degree of amplification on the discrepancy between two distributions. 
The proposed framework operates as a self-corrective decoding mechanism that is specifically designed to trick AI-text detectors. By dynamically identifying and penalizing machine-preferred token preferences, CoPA effectively reduces entrenched linguistic biases and enables the generation of sentences with more expressiveness and lexical diversity, achieving an enhanced capability to fool LLM-generated text detectors.  

\textbf{Reduce machine styles by amplifying them.}  
Previous studies have shown that rewriting machine texts using another LLM can reduce some machine-specific features, although not sufficiently for launching effective attacks~\cite{sadasivan2023can}. This is because the rewritten texts inherit a mixture of stylistic and lexical preferences from different LLMs, thus increasing the text diversity.  
\begin{figure}
    \centering
    \includegraphics[width=0.93\linewidth]{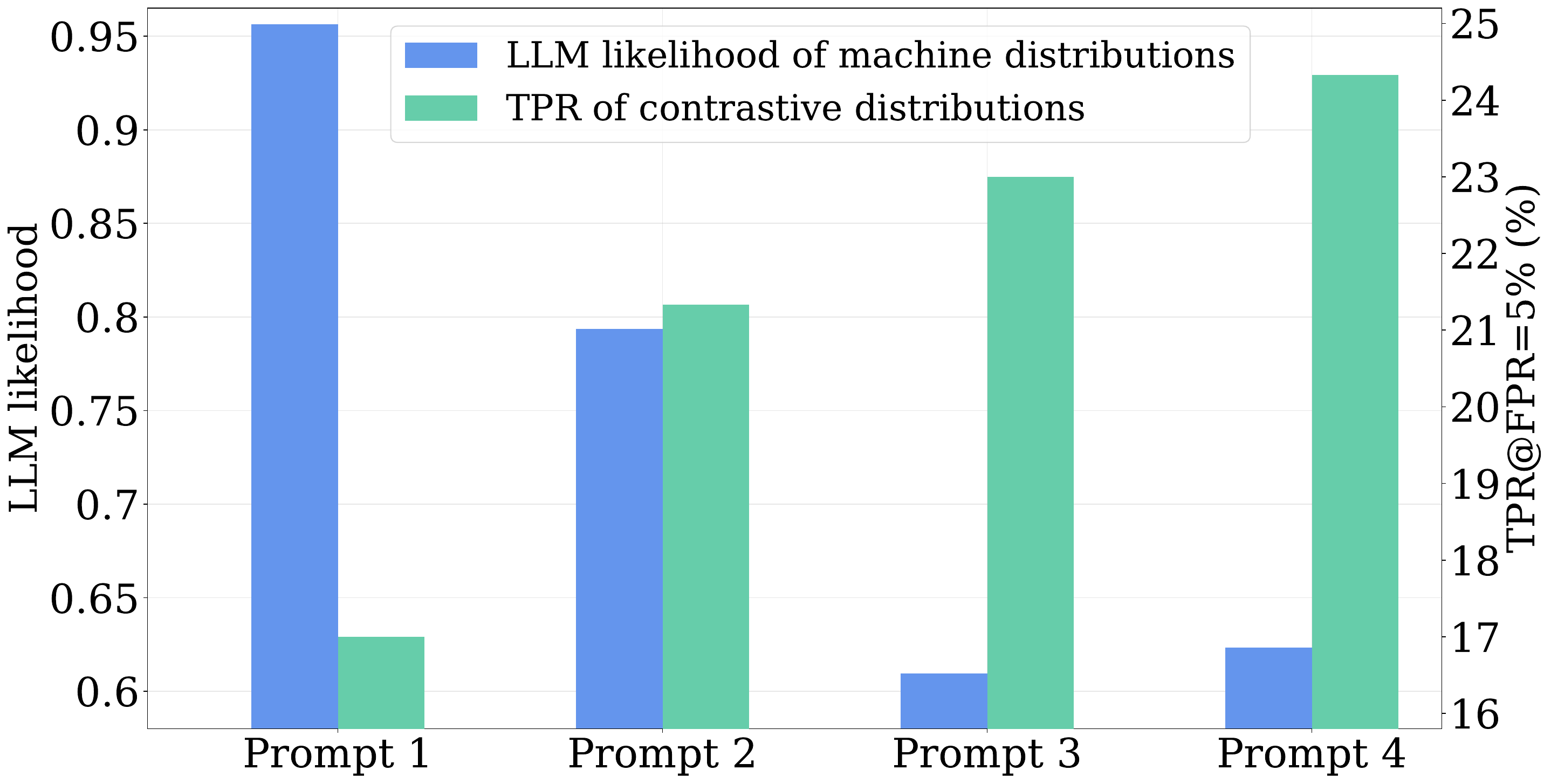}
    \caption{Fast-DetectGPT detected LLM likelihood of texts from different machine distributions $p_m$ induced by various machine prompts $x_m$. We also present the detection TPR of their corresponding contrastive distributions $p_c$. See Appendix \ref{sec:prompt_analysis} for details of used prompts.}
    \label{fig:reduce_by_amplify}
    \vspace{-0.5em}
\end{figure}
However, in our contrastive design, the machine-style distribution \( p_m \) actually serves as a negative reference to be subtracted from the human distribution. 
Employing a regular paraphrasing prompt to obtain $p_m$ may inadvertently dilute its machine-specific characteristics and weaken the effectiveness of the contrastive operation. 

A reasonable solution involves constructing $p_m$ as a highly salient and concentrated machine-style token distribution, wherein high-probability tokens are strongly machine-related and more likely to trigger detection. 
We achieve this by identifying a prompt $x_m$ that can amplify the LLM likelihood of generated sentences.
Fig. \ref{fig:reduce_by_amplify} empirically validates our strategy, \ie magnifying machine-style features in $p_m$ can, in turn, promote a refined distribution that more closely resembles authentic human writing, further enhancing the attack effectiveness.

\textbf{Adaptive Truncation for plausibility.} Another important issue is that CoPA utilizes the whole token distribution to measure the difference. However, there may be occasions where certain high-probability tokens overlap between \( p_h' \) and \( p_m \). The subtraction operation may penalize the probabilities of reasonable and valid tokens while rewarding casual and unrelated ones \cite{fang2025grounding}, thus compromising the coherence and semantic consistency of generated sentences.  
To address this, we incorporate a token constraint mechanism \cite{li2023contrastive}, which applies an adaptive pruning to the output tokens:  
\begin{equation}
\begin{split}
y_t\sim p_{c}(\cdot|x_h, x_m, &y_m, y_{<t}), \text{ s.t. } y_t \in \mathcal{V}_{top}(y_{<t}), \\
\quad\mathcal{V}_{top}(y_{<t})=\Big\{y_t&\in\ \mathcal{V}:p_h'( y_t|x_h, y_m, y_{<t}) \\ 
&\geq \alpha \max_v p_h'(v|x_h, y_m, y_{<t})\Big\},
\label{eq:LLM_constrain_generation}
\end{split}
\end{equation}
where $\mathcal{V}$ denotes the vocabulary set of $f_{\theta}$ and $\alpha$ is the hyperparameter to adjust clipping. By introducing this adaptive pruning mechanism, CoPA leverages the confidence scores of the human-like distribution to refine the contrastive distribution, which restricts the decision-making to a more reliable token candidate pool and suppresses the selection of unsuitable tokens. 

\subsection{Theoretical Analysis}

Apart from empirical analysis, we build a theoretical framework to confirm the superiority of CoPA in simulating authentic human writing. As previously stated, \( p_{h} \) denotes the real human-chosen word distribution, \( p_{h}' \) and \( p_{m} \) represent human-like and machine-like token distributions elicited from the LLM using prompts $x_h$ and $x_m$, respectively. The objective is to prove that the distribution \( p_{c} \), derived by contrasting $p_h'$ and $p_m$, aligns more closely with the human preferences \( p_{h} \). 

To mathematically measure the difference between distributions $p_h$ and $p_c$, we first introduce an auxiliary function based on the KL divergence. 
\begin{definition}[Auxiliary Distance Function] 
Let $\mathbb{KL}$ denotes the KL divergence, the distributional distance between $p_h$ and $p_c$ is a unary function of $\lambda$, which is characterized as
\begin{equation}
\begin{split}
g(\lambda)\coloneqq\mathbb{KL}(p_{h}||(1+\lambda)p_{h}'- \lambda p_{m}).
\end{split}
\end{equation}
\end{definition}
$g(\lambda)$ inherit several good properties from the KL divergence, based on which we derive the critical Theorem that guarantees the effectiveness of CoPA:
\begin{proposition}
$g(\lambda)$ is a convex function. If $g(\lambda)$ is not constant, it has a unique minimum point $\lambda_{*}$. 
\label{prop:unique}
\end{proposition}
\begin{theorem}
\label{theorem:conclusion}
If $g'(0)<0$, then $\lambda_{*}>0$ and for any $\lambda\in(0,\lambda_{*}]$, we have
\begin{equation}
\begin{split}
\mathbb{KL}(p_{h}||(1+\lambda)p_{h}'-\lambda p_{m})<\mathbb{KL}(p_{h}||p_{h}').
\end{split}
\end{equation}
\end{theorem}
The detailed proofs of Proposition \ref{prop:unique} and Theorem \ref{theorem:conclusion} are provided in Appendix \ref{sec:proof}.
Theorem \ref{theorem:conclusion} reveals that by adequately selecting $\lambda$,
CoPA drives the resultant distribution $p_c$ closer to the authentic human distribution $p_h$ than the human-like distribution $p_h'$, which is directly elicited from the LLM using $x_{h}$. This theoretically validates the necessity and effectiveness of our contrastive strategy.

\begin{figure}
    \centering
    \includegraphics[width=0.6\linewidth]{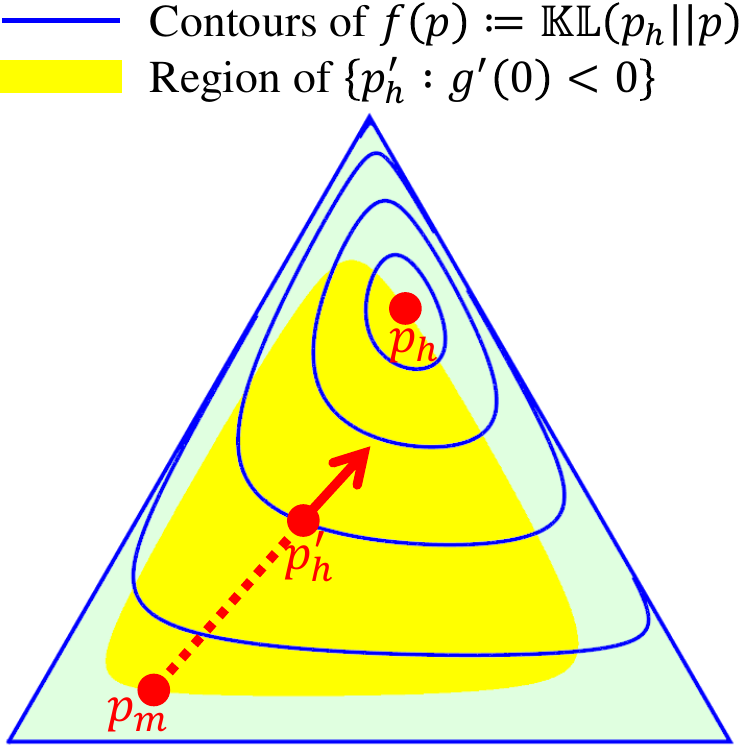}
    \caption{Illustration of the premise of \cref{theorem:conclusion}. Let $|\mathcal{V}|=3$ and thus $\mathbb{P}^{\mathcal{V}}$ is a triangle. We draw the contours of $f(p)\coloneqq\mathbb{KL}(p_{h}||p)$. The closer to $p_{h}$, the lower the KL divergence with $p_{h}$. If $p_{h}'-p_{m}$ points to the inside of the contour at $p_{h}'$, then $f(p)$ decreases at $p_{h}'$ along $p_{h}'-p_{m}$, \ie $g(\lambda)$ decreases at $\lambda=0$. In this case $g'(0)<0$ is satisfied and \cref{theorem:conclusion} is applicable. In practice $p_{h}'$ is usually between $p_{m}$ and $p_{h}$, so $g'(0)<0$ is usually satisfied and \cref{theorem:conclusion} is generally applicable.}
    \label{fig:ill4theorem1}
\end{figure}

Note that the premise of \cref{theorem:conclusion} is $g'(0)<0$. 
We use \( |\mathcal{V}| = 3 \) as an example to illustrate its rationality. 
As in Figure \ref{fig:ill4theorem1}, we calculate the area wherein probability distributions satisfy \( g(0)'<0 \). 
In essence, \( p_h' \) is generated by the LLM and hence is constrained by the inherent language priors, which prevent it from deviating significantly from the machine-featured distribution \( p_m \). Meanwhile, we use a carefully crafted human-like prompt to guide \( p_h' \) to move from \( p_m \) towards \( p_h \). As a result, \( p_h' \) typically falls within the region that satisfies \( g'(0) < 0 \). Therefore, this premise generally holds and thus \cref{theorem:conclusion} is applicable in practice, which is further confirmed by experimental results in Sec. \ref{sec:main_exp}.
Based on LLM's prediction paradigm, we contrast the output logits in practice, achieving excellent performance in misleading detection models. We also note that researchers should examine the validity of this assumption in their specific setting before applying our theoretical framework.

\section{Experiments}
\begin{table*}[htbp]
  \centering
  \setlength{\tabcolsep}{4.07pt}
  \caption{Comparison of different paraphrasing attacks against 8 text-detection algorithms (at 5\% FPR) using GPT-3.5-turbo generated texts from three different datasets. The best performances are bolded.}
  \vspace{-0.5em}
    \resizebox{\linewidth}{!}{\begin{tabular}{cccccccccccr} \toprule
    \multirow{2}[0]{*}{Dataset} & \multirow{2}[0]{*}{Attack} & \multirow{2}[0]{*}{Sim} & \multicolumn{8}{c}{Defense}                                & \multicolumn{1}{c}{\multirow{2}[0]{*}{Avg.}} \\  \cmidrule(lr){4-11} 
          &       &       & LogRank & DetectGPT & DNA-GPT & Fast-DetectGPT & Raidar & TOCSIN & RoBERTa & R-Detect &  \\ \midrule
    \multirow{4}[0]{*}{XSum} & No Attack & -     & 63.33 & 26.67 & 80.00 & 95.33 & 28.75 & 98.00 & 66.67 & 69.67 & 66.05 \\
          & Dipper & 86.67 & 15.67 & \textbf{2.67} & 27.33 & 76.33 & 13.75 & 74.67 & 86.67 & 48.00 & 43.14 \\
          & Raidar & 100.00 & 49.00 & 12.00 & 22.64 & 84.67 & 16.25 & 90.67 & 55.33 & 68.67 & 49.90 \\
          & \gc Ours  & \gc  94.00 & \gc  \textbf{4.67} & \gc  4.00  & \gc \textbf{21.33} & \gc \textbf{17.00} & \gc \textbf{0.00} & \gc \textbf{26.67} & \gc \textbf{22.67} & \gc \textbf{4.67} & \gc 
 \textbf{12.63} \\ \midrule
    \multirow{4}[0]{*}{SQuAD} & No Attack & -     & 67.00 & 12.67 & 41.33 & 93.50 & 22.00 & 92.67 & 41.33 & 79.67 & 56.27 \\
          & Dipper & 75.33 & 23.67 & 2.67  & 10.00 & 77.67 & 5.00  & 75.33 & 67.33 & 53.33 & 39.38 \\
          & Raidar & 95.33 & 58.00 & 12.33 & 19.33 & 81.67 & 26.00 & 84.00 & 32.67 & 73.50 & 48.44 \\
          & \gc Ours  & \gc 88.67 & \gc \textbf{8.33} & \gc \textbf{2.67} & \gc \textbf{8.67} & \gc \textbf{27.50} & \gc \textbf{5.00} & \gc \textbf{25.33} & \gc \textbf{7.33} & \gc \textbf{3.67} & \gc \textbf{11.06} \\ \midrule
    \multirow{4}[0]{*}{LongQA} & No Attack & -     & 73.83 & 33.33 & 10.67 & 86.00 & 36.25 & 88.67 & 38.67 & 89.33 & 57.09 \\
          & Dipper & 94.67 & 28.83 & 6.00  & 0.67  & 75.00 & 5.00  & 67.33 & 65.33 & 74.00 & 40.27 \\
          & Raidar & 100.00 & 59.33 & 22.67 & 1.33  & 72.67 & 33.75 & 77.33 & 28.00 & 79.50 & 46.82 \\
          & \gc Ours  & \gc 95.33 & \gc \textbf{14.50} & \gc \textbf{5.00} & \gc \textbf{0.00} & \gc \textbf{11.33} & \gc \textbf{0.00} & \gc \textbf{16.00} & \gc \textbf{6.67} & \gc \textbf{6.00} & \gc \gc \textbf{7.44} \\ \bottomrule
    \end{tabular}}
    \vspace{-0.3em}
  \label{tab:main}%
\end{table*}%

\begin{figure*}[t]
\begin{center}
\includegraphics[width=\linewidth]{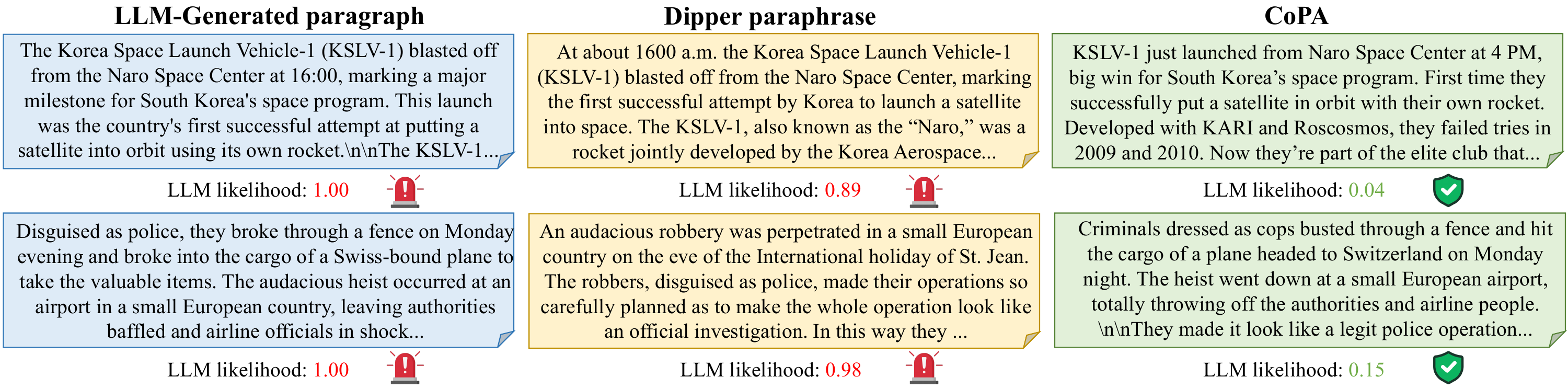}
\end{center}
\vspace{-0.8em}
\caption{Paraphrased sentences by Dipper and our CoPA. We use Fast-DetectGPT to provide the LLM likelihood.}
\label{fig:visual}
\vspace{-0.3em}
\end{figure*}

\subsection{Experimental Setup}
\textbf{Datasets.}
We evaluate on three widely adopted datasets spanning various linguistic styles and content, including (1) XSum for news articles \cite{narayan2018don}, (2) SQuAD for Wikipedia contexts \cite{rajpurkar2016squad}, and (3) LongQA for long-form question answering, where LLM answers a how/why question within 250-350 words \cite{fan2019eli5}. We follow \cite{baofast} and randomly select 150 samples for evaluation. 

\textbf{Baselines.}
We compare our method with the state-of-the-art (SOTA) surrogate-free paraphrase attack Dipper \cite{krishna2023paraphrasing}. We adopt the setup of 60 lexical diversity and 60 order diversity for Dipper to achieve its best performance.
Additionally, we reproduce the attack introduced in \cite{maoraidar}, which leverages an LLM with the query \textit{``Help me rephrase it in human style.''} to rewrite machine texts (denoted as Raidar-A). For fairness, we use the same LLM for both our paraphraser and baselines. Note that we also reveal the superiority of CoPA over \cite{shi2024red} that relies on an extra surrogate model in \cref{sec:add_results}.

For detection algorithms, we consider diverse methods including training-free LogRank \cite{solaiman2019release}, DetectGPT \cite{mitchell2023detectgpt}, DNA-GPT \cite{yangdna}, Fast-DetectGPT \cite{baofast}, Raidar \cite{maoraidar}, TOCSIN \cite{ma2024zero}, and training-based  RoBERTa \cite{liu2019roberta} provided by OpenAI and the R-detect \cite{songdeep}.

\textbf{Metrics.}
We analyze the performance using two key metrics.
(1) Detection accuracy. In real-world applications, it is crucial to guarantee that human-written text should almost never be misclassified as machine-generated \cite{krishna2023paraphrasing}, \ie satisfying a very low false positive rate (FPR). Hence, we follow Dipper and report the true positive rate (TPR) at a fixed FPR. Specifically, we set a relatively high FPR of 5\% to significantly reveal the performance improvements.
Please refer to Appendix \ref{sec:add_results} for results with FPR=1\%. 
(2) Semantic similarity. The rewritten sentences should preserve the original semantics. Similar to Dipper, we employ the P-SP \cite{wieting2022paraphrastic} model, a specialized embedding model trained on a filtered paraphrase corpus \cite{wieting2018paranmt}, to measure the semantic discrepancy. We align with Dipper and consider the semantics being preserved if the P-SP score exceeds the average real-human paraphrase score of 0.76. Moreover, we provide \textit{more analysis including text perplexity, GPT-4 assisted and human evaluation in \cref{sec:text_quality}}.

\textbf{Implementation Details.}
For attack hyperparameters, we set the contrast intensity \(\lambda = 0.5\) and the clipping factor \(\alpha=1\text{e}^{-5}\). 
Unless stated otherwise, we employ a single paraphrasing iteration. We employ Qwen2.5-72B-Instruct \cite{qwen2.5} as the paraphraser. Due to page limits, we provide results paraphrased by more LLMs in \cref{sec:add_results}. More details about the human and machine-like prompts are in Appendix \ref{sec:exp_details}.

\subsection{Performance Evaluation}
\label{sec:main_exp}
We test machine texts generated by GPT-3.5-turbo and present results on three datasets in Table \ref{tab:main}. 

\begin{table*}[htbp]
  \centering
  \caption{Attack Performance (at 5\% FPR) of texts generated by more source LLMs based on XSum dataset.}
  \vspace{-0.5em}
    \setlength{\tabcolsep}{4.07pt}
    \resizebox{\linewidth}{!}{\begin{tabular}{cccccccccccr} \toprule
    \multirow{2}[0]{*}{Model} & \multirow{2}[0]{*}{Attack} & \multirow{2}[0]{*}{Sim} & \multicolumn{8}{c}{Defense}                                   & \multicolumn{1}{c}{\multirow{2}[0]{*}{Avg.}} \\ \cmidrule(lr){4-11}
          &       &       & LogRank & DetectGPT & DNA-GPT & Fast-DetectGPT & Raidar & TOCSIN & RoBERTa & R-Detect &  \\ \midrule
    \multirow{4}[0]{*}{GPT-4} & No Attack & -     & 30.00 & 6.00  & 35.33 & 51.67 & 24.17 & 73.33 & 32.67 & 46.00 & 37.40 \\
          & Dipper & 91.33 & 8.67  & 0.67  & 30.67 & 64.33 & 20.83 & 64.67 & 78.00 & 37.67 & 38.19 \\
          & Raidar & 100.00 & 35.00 & 9.50  & 34.67 & 68.33 & 20.83 & 82.00 & 47.33 & 60.50 & 44.77 \\
          & \gc Ours  & \gc 94.67 & \gc \textbf{2.00} & \gc \textbf{0.67} & \gc \textbf{18.67} & \gc \textbf{15.33} & \gc \textbf{10.83} & \gc \textbf{20.00} & \gc \textbf{20.67} & \gc \textbf{6.83} & \gc \textbf{11.88} \\ \midrule
    \multirow{4}[0]{*}{Claude 3.5} & No Attack & -     & 42.67 & 21.67 & 24.67 & 50.00 & 36.67 & 70.00 & 19.33 & 30.67 & 36.96 \\
          & Dipper & 82.00 & 18.17 & \textbf{0.33} & 20.67 & 38.67 & 0.00  & 36.67 & 77.33 & 40.33 & 29.02 \\
          & Raidar & 100.00 & 46.83 & 18.77 & 15.33 & 41.33 & 41.64 & 66.00 & 20.67 & 38.00 & 36.07 \\
          & \gc Ours  & \gc 98.00 & \gc \textbf{1.33} & \gc 0.67  & \gc \textbf{6.67} & \gc \textbf{4.00} & \gc \textbf{0.00} & \gc \textbf{8.00} & \gc \textbf{7.33} & \gc \textbf{1.33} & \gc \textbf{3.67}  \\ \bottomrule
    \end{tabular}}
  \label{tab:more_source}%
  \vspace{-0.2em}
\end{table*}%

\textbf{Attack Effectiveness.} By conducting a self-introspective correction on token distributions, CoPA remarkably enhances the attack over baseline attacks, 
\eg an average improvement of 30.55\% in fooling text detectors across three datasets. Although Dipper demonstrates satisfactory performance against several detectors, it becomes significantly less effective when facing more advanced algorithms such as FastDetectGPT. In contrast, our method consistently exhibits impressive attack efficacy across various text detectors.  
Notably, while Raidar-A and our method employ the same LLM as the paraphraser, CoPA greatly outperforms Raidar-A, validating the effectiveness of our designed prompt and contrastive paraphrasing mechanism.

As for the text quality of rewritten sentences, we demonstrate that CoPA achieves an average semantic similarity score exceeding 90\% across various datasets, confirming that our method effectively preserves semantic fidelity during rewriting. 
While Raidar-A exhibits greater text similarity, its attack effectiveness remains considerably limited. As a comparison, CoPA achieves both excellent attack effectiveness and semantic consistency.

\textbf{Visualization of Rewritten Texts.} We provide examples of rewritten texts before and after the paraphrasing in Figure \ref{fig:visual}. As expected, the rewritten sentences maintain semantic consistency while exhibiting richer and diverse human-like expressions. This underpins our success in fooling text detectors and presents a practical attack method.

\subsection{Attacks on More Source LLMs.} In addition to GPT-3.5-turbo, we also consider the machine texts generated by recently prevalent LLMs, including GPT-4 \cite{achiam2023gpt} and Claude-3.5 \cite{antrophic}. Besides, we provide results on GPT-4o \cite{achiam2023gpt} and Gemini-1.5 Pro \cite{team2024gemini} in Appendix \ref{exp:more_source}. 

Table \ref{tab:more_source} demonstrates that our method continues to achieve excellent attack efficacy and semantic similarity across various LLM-generated texts, greatly outperforming the SOTA method Dipper. For detection of GPT-4 texts under Fast-DetectGPT, Dipper even increases the likelihood to be classified as machine-generated than the \textit{No Attack} baseline.
In contrast, CoPA stably achieves outstanding fooling rates across various source models, confirming the robustness of the proposed contrastive paraphrase.
Another observation is that the detection performance on clean texts generated by more recent models is significantly reduced. The decline may stem from the ability of advanced LLMs to generate varying sentences that are more challenging to detect, underscoring the urgent need for more reliable detection systems.
\begin{figure}[t]
\begin{center}
\includegraphics[width=\linewidth]{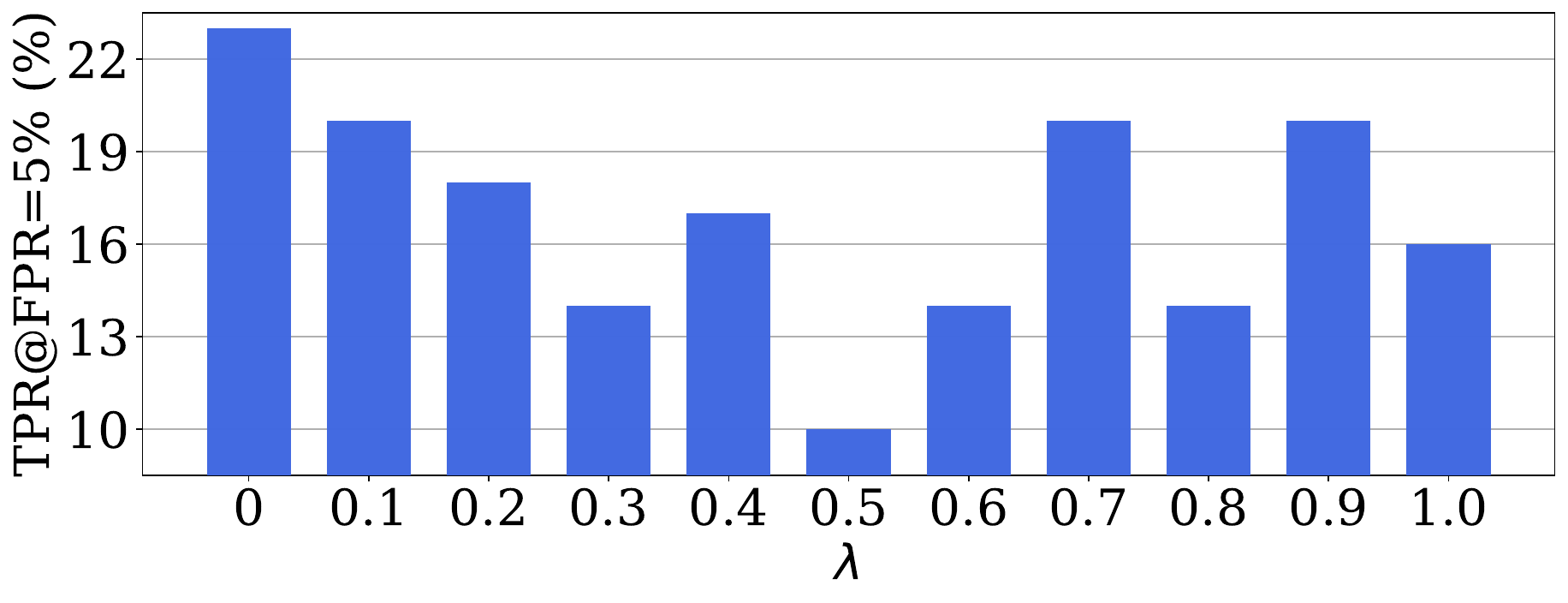}
\end{center}
\vspace{-1.2em}
\caption{Comparison of detection accuracy on the first 50 samples from XSum under different values of $\lambda$ against Fast-DetectGPT \cite{baofast}.}
\label{fig:lambda}
\vspace{-0.8em}
\end{figure}
\subsection{Ablation Study}
\label{sec:ablation}
We then investigate the effect of different factors. More ablation studies are in Appendix \ref{exp:sample_params}.

\textbf{Impact of contrastive coefficient \(\lambda\).}  
During decoding, the hyperparameter \(\lambda\) serves as a critical regulation factor for the contrast strength. 
As shown in Figure \ref{fig:lambda}, positive values of $\lambda$ consistently enhance the fooling rates relative to $\lambda=0$, again confirming the effectiveness of our contrastive paraphrasing mechanism. Note that CoPA attains optimal performance at \(\lambda = 0.5\), which is then chosen as the default setting for our experiments. Notably, the general trend of TPR over the $\lambda$ roughly aligns with our preceding theoretical analysis, verifying the rationality of our established theory.

\begin{figure}[t]
\begin{center}
\includegraphics[width=\linewidth]{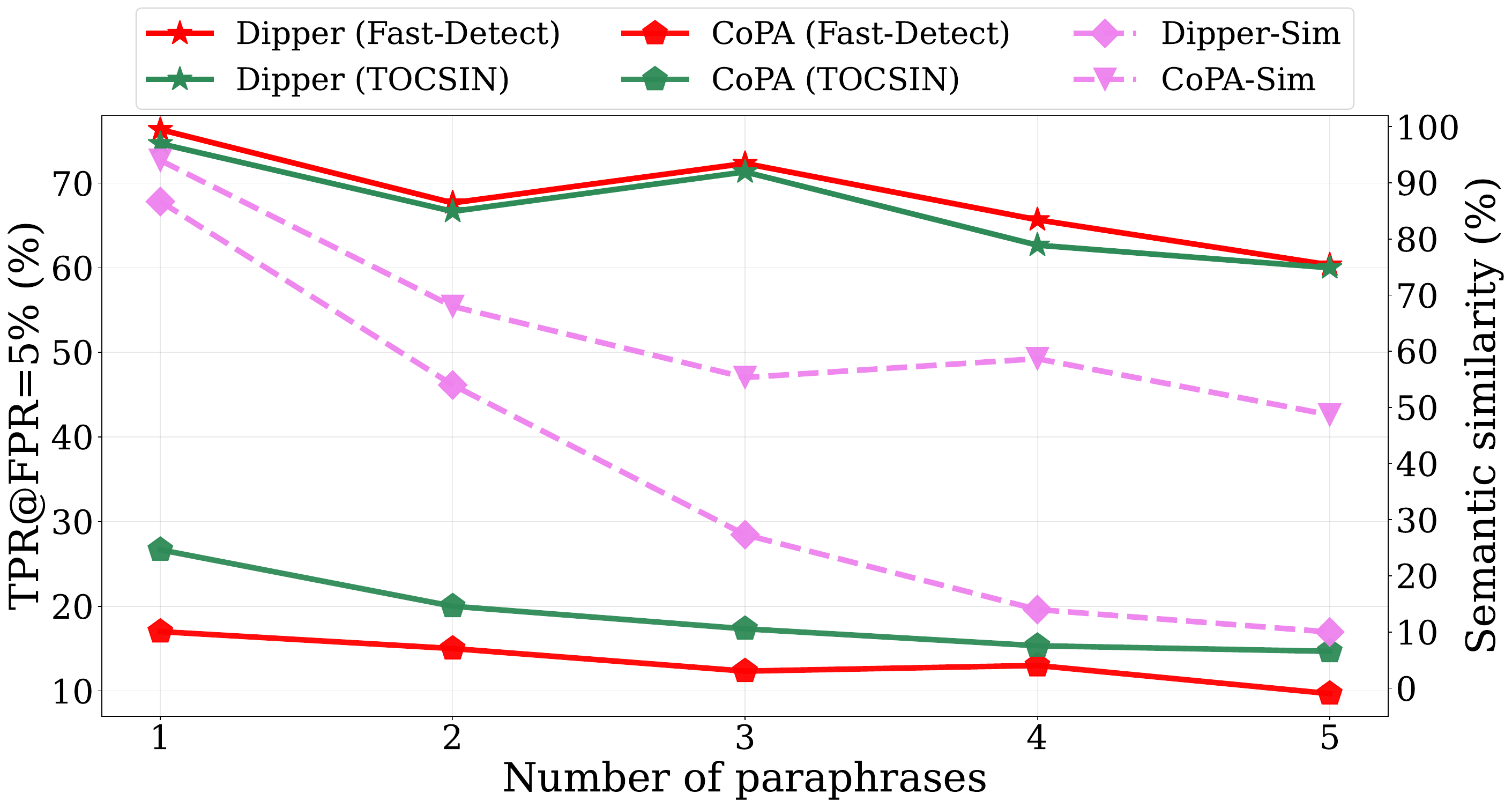}
\end{center}
\vspace{-0.8em}
\caption{Performance under different numbers of paraphrases against Fast-DetectGPT \cite{baofast} and TOCSIN \cite{ma2024zero} on samples generated by GPT-3.5-turbo from the XSum dataset. The dashed lines describe semantic similarity.}
\vspace{-0.8em}
\label{fig:multiple_paraphrases}
\end{figure}

\textbf{Impact of Multiple Paraphrases.} 
Each LLM-generated text is rewritten only once in our former experiments. We further analyze the influence of multiple rewrites on the results. As shown in Figure \ref{fig:multiple_paraphrases}, increasing the number of rewrites generally strengthens attack effectiveness. However, the performance gain is limited against two advanced defenses, and the semantic similarity of Dipper-rewritten texts sharply drops as the iterations increase. These factors reduce the utility of using multiple paraphrases to improve the attack \cite{sadasivan2023can}. Also, this again highlights the superiority of our method, which achieves outstanding fooling rates via only a single paraphrasing.


\section{Conclusion}
This paper proposes CoPA, a simple yet highly effective paraphrasing attack against AI-generated text detectors. CoPA constructs a machine-style token distribution as a negative contrast for reducing linguistic biases of LLMs and facilitating the generation of richer and more diverse sentences.  
Through both theoretical analysis and experimental validation, we fully demonstrate the superiority of the proposed method across various scenarios. We envision CoPA as a powerful tool for auditing the robustness of detection systems, inspiring future development of more robust detection algorithms.
\section*{Limitations}
While our method avoids the overhead of training a dedicated paraphraser by leveraging an off-the-shelf LLM, the contrastive paraphrasing mechanism requires two forward passes to construct the contrastive token distribution, bringing additional latency during next-token prediction. This may limit the practicality of the proposed attack in real-time applications. Moreover, using off-the-shelf LLMs to paraphrase texts can lead to the semantics of output texts deviating from the original sentence, which may require multiple generations to maintain the semantic similarity. Besides, although human evaluation results in Appendix~\ref{sec:text_quality} indicate that CoPA-paraphrased texts are preferred over those from Dipper, we did not systematically account for the detailed linguistic backgrounds of evaluators and may introduce bias. A more comprehensive human study is needed to validate the general quality of the output sentences. Finally, this work follows prior studies and focuses exclusively on English text. Extending the contrastive paraphrasing framework to other languages, such as Chinese and Spanish, would be valuable for its broader applicability.
\section*{Ethical Statement}
This paper presents a novel method aimed to advance the research field of LLM-generated text detection. Note that all experiments are conducted within controlled laboratory environments. We do not expect the proposed method to serve as a powerful tool for potential adversaries but to raise society's broader awareness of the vulnerability of current AI-text detectors.
Also, the exceptional attack performance highlights the practical limitations of current detectors. Researchers of the open-source community are encouraged to conduct stress tests on their detectors against the proposed attack, based on which future studies can develop more robust stronger detectors. \textbf{Furthermore, we conduct a preliminary study to alleviate the proposed threat via an adaptive defense that adversarially trains a Roberta-based detector using texts paraphrased by CoPA in \cref{sec:adpative_defense}.} 

All the codes, models, and datasets used in this study are consistent with their intended use and comply with the MIT License. To promote further research, we will open-source our paraphrasing tool along with the related code, model, and data. 

\section*{Acknowledgement}
This work is supported in part by the National Natural Science Foundation of China under grant 62171248, 62301189, 62576122, and Shenzhen Science and Technology Program under Grant KJZD20240903103702004, JCYJ20220818101012025, GXWD20220811172936001.

\bibliography{custom}
\newpage
\appendix
\appendix
\section{Theorem and Proof}
\label{sec:proof}
\setcounter{definition}{0}
\begin{definition}
$g(\lambda)\coloneqq\mathbb{KL}(p_{h}||(1+\lambda)p_{h}'-\lambda p_{m})$.
\end{definition}

\begin{proposition}
$g(\lambda)$ is a convex function.
\end{proposition}
\begin{proof}
$\mathbb{KL}$ is convex and $g(\lambda)$ is the restriction of $\mathbb{KL}$ on a line, so $g(\lambda)$ is also convex.
\end{proof}

\begin{proposition}
\label{定理：存在唯一最小值点}
If $g(\lambda)$ is not constant, then $g(\lambda)$ has a unique minimum point.
\end{proposition}
\begin{proof}
The non-constancy of $g(\lambda)$ implies that $p_{h}'\neq p_{m}$. The domain of $g(\lambda)$ is
\begin{align*}
I\coloneqq\bigcap_{v\in\mathcal{V}}\{\lambda\in\mathbb{R}:0\leq (1+\lambda)p_{h}'^{(v)}-\lambda p_{m}^{(v)}\leq 1\}.
\end{align*}

$[-1,0]\subseteq I$ so $I$ is non-empty. $I$ is the intersection of some closed intervals, so $I$ is also a closed interval. Note that $g(\lambda)$ is continuous, so $g(\lambda)$ has a minimum value.

Assume that $g(\lambda)$ has minimum points $\lambda_1$ and $\lambda_2$ with $\lambda_1<\lambda_2$. By the convexity, $g(\lambda)$ is constant on $[\lambda_1,\lambda_2]$. Note that $g(\lambda)$ is an analytic function. By Corollary 1.2.6 in \cite{krantz2002primer}, the constancy of $g(\lambda)$ on $[\lambda_1,\lambda_2]$ implies the constancy on $I$. This contradicts that $g(\lambda)$ is not constant, so the minimum point of $g(\lambda)$ is unique.
\end{proof}

\begin{remark}
Corollary 1.2.6 in \cite{krantz2002primer} applies to open intervals, and $g(\lambda)$ is continuous at the endpoints, so it also applies to our closed intervals, $[\lambda_1,\lambda_2]$ and $I$.
\end{remark}

\begin{definition}
The minimum point of $g(\lambda)$ is $\lambda_*$.
\end{definition}

\setcounter{theorem}{0}
\begin{theorem}
\label{定理：在区间内下降}
If $g'(0)<0$, then $\lambda_*>0$ and for any $\lambda\in(0,\lambda_{*}]$, we have
\begin{align}
\label{公式：KL散度变小}
\mathbb{KL}(p_{h}||(1+\lambda)p_{h}'-\lambda p_{m})<\mathbb{KL}(p_{h}||p_{h}').
\end{align}
\end{theorem}
\begin{proof}
$g'(0)<0$ implies that $g(\lambda)$ is not constant, so $\lambda_*$ is well-defined by Proposition~\ref{定理：存在唯一最小值点}.

$g'(0)\neq0$ implies $\lambda_*\neq0$. Assume $\lambda_*<0$. By the first-order condition of convex functions,
\begin{align*}
g(\lambda_{*})\geq g(0)+g'(0)\lambda_{*}>g(0).
\end{align*}
This contradicts that $g(\lambda_*)$ is minimum, so $\lambda_*>0$.

By the definition of convex functions, we have
\begin{align*}
g(\lambda)&\leq\frac{\lambda}{\lambda_{*}}g(\lambda_{*})+\left(1-\frac{\lambda}{\lambda_{*}}\right)g(0),\\
&=g(0)-\frac{\lambda}{\lambda_{*}}\left(g(0)-g(\lambda_{*})\right),\\
&<g(0)
\end{align*}
for any $\lambda\in(0,\lambda_{*}]$. $g(\lambda)<g(0)$ is just (\ref{公式：KL散度变小}).
\end{proof}

\section{Experimental Details}
\label{sec:exp_details}
We follow the implementation of \cite{wang2024raft,baofast} to generate machine texts for XSum and SQuAD while adopting the approach of Dipper \cite{krishna2023paraphrasing} for LongQA. To reproduce DetectGPT and Fast-DetectGPT, we align with \cite{wang2024raft} and employ GPT2-XL \cite{radford2019language} as the surrogate model to generate samples. As for DNA-GPT \cite{yangdna}, we adopt the basic setup surrogated on GPT-3.5-turbo.
For the LLM decoding, our experiments adopt the default setup of sampling parameters, \ie no probability clipping and \(T = 1\). We run all experiments in NVIDIA RTX A6000 GPUs. 

Below, we provide our carefully designed prompts to elicit human-like and machine-like token distributions from the off-the-shelf LLM.

(1) For human-style distributions: 
An intuitive way is to construct prompts that directly ask LLMs to produce human-like sentences by encouraging vivid lexical substitution and diverse sentence structures.
However, compared to a direct request for human-toned texts, we empirically find that framing a realistic human conversation scene to LLMs leads to more vivid and diverse generations, strengthening the attack effectiveness.
\begin{tcolorbox}[ title=System Prompt]
You are a helpful paraphraser. You are given an input passage 'INPUT'. You should paraphrase 'INPUT' to print 'OUTPUT'. 'OUTPUT' should preserve the meaning and content of 'INPUT'. 'OUTPUT' should not be very shorter than 'INPUT'.
\end{tcolorbox}
\begin{table*}[htbp]
  \centering
  \caption{Detection accuracy (at 5\% FPR) of texts generated by GPT-4o and Gemini-1.5-Pro based on XSum dataset.}
  \vspace{-0.5em}
    \setlength{\tabcolsep}{4.07pt}
    \resizebox{\linewidth}{!}{\begin{tabular}{cccccccccccr} \toprule
    \multirow{2}[0]{*}{Model} & \multirow{2}[0]{*}{Attack} & \multirow{2}[0]{*}{Sim} & \multicolumn{8}{c}{Defense}                                   & \multicolumn{1}{c}{\multirow{2}[0]{*}{Avg.}} \\ \cmidrule(lr){4-11}
          &       &       & LogRank & DetectGPT & DNA-GPT & Fast-DetectGPT & Raidar & TOCSIN & RoBERTa & R-Detect &  \\ \midrule
    \multirow{4}[0]{*}{GPT-4o} & No Attack & -     & 33.33 & 1.83  & 37.33 & 12.00 & 100.00 & 24.00 & \textbf{3.33} & 26.67 & 29.81 \\
          & Dipper & 77.33 & 16.00 & 1.33  & 39.33 & 42.33 & 11.33 & 42.67 & 68.00 & 43.67 & 33.08 \\
          & Raidar & 99.33 & 24.00 & 11.67 & 42.67 & 46.33 & 5.00  & 64.67 & 15.33 & 56.67 & 33.29 \\
          & \gc Ours  &\gc 96.67 & \gc \textbf{3.00} & \gc \textbf{0.67} & \gc \textbf{22.00} & \gc \textbf{6.33} & \gc \textbf{5.00} & \gc \textbf{10.00} & \gc 16.67 &\gc  \textbf{4.67} &\gc  \textbf{8.54} \\ \midrule
    \multirow{4}[0]{*}{Gemini-1.5-Pro} & No Attack & -     & 21.67 & 13.00 & 36.00 & 32.00 & 28.00 & 33.33 & 12.67 & 24.67 & 25.17 \\
          & Dipper & 80.67 & 11.00 & 0.67  & 23.33 & 51.17 & 5.00  & 48.00 & 72.67 & 35.67 & 30.94 \\
          & Raidar & 100.00 & 31.00 & 8.67  & 26.00 & 39.83 & 30.00 & 48.67 & 22.67 & 42.67 & 31.19 \\
          & \gc Ours  &\gc 93.33 & \gc \textbf{2.00} & \gc\textbf{2.67} & \gc\textbf{6.67} & \gc\textbf{10.00} & \gc\textbf{2.00} & \gc \textbf{10.00} & \gc \textbf{9.33} & \gc\textbf{2.00} & \gc\textbf{5.58} \\ \bottomrule
    \end{tabular}}
  \label{tab:exp_more_more_source}%
  
\end{table*}%

\begin{table*}[htbp]
  \centering
  \setlength{\tabcolsep}{4.07pt}
  \caption{Comparison of different paraphrasing attacks against 8 text-detection algorithms (at 1\% FPR) using GPT-3.5-turbo generated texts from three different datasets. The best performances are bolded.}
    \resizebox{\linewidth}{!}{\begin{tabular}{cccccccccccc} \toprule
    \multirow{2}[0]{*}{Dataset} & \multirow{2}[0]{*}{Attack} & \multirow{2}[0]{*}{Sim} & \multicolumn{8}{c}{Defense}                                & \multicolumn{1}{c}{\multirow{2}[0]{*}{Avg.}} \\  \cmidrule(lr){4-11} 
          &       &       & LogRank & DetectGPT & DNA-GPT & Fast-DetectGPT & Raidar & TOCSIN & RoBERTa & R-Detect &  \\ \midrule
    \multirow{4}[0]{*}{XSum} & No Attack & - & 32.44 & 6.00  & 41.33 & 83.00 & 5.75  & 94.67 & 44.67 & 49.33 & 44.65 \\
          & Dipper & 86.67 & 7.33  & \textbf{0.00}  & 8.00  & 44.67 & 2.75  & 46.00 & 74.00 & 28.33 & 26.39 \\
          & Raidar & 100.00 & 21.00&	1.00&	5.33&	57.33&	3.25&	85.33&	39.33&	58.67&	33.91 \\
          & \gc Ours  & \gc 94.00 & \gc \textbf{2.00} & \gc 	0.01& \gc 	\textbf{5.33}& \gc 	\textbf{4.67}& \gc 	\textbf{0.00}& \gc 	\textbf{18.00}& \gc 	\textbf{9.33}& \gc 	\textbf{0.00}& \gc 	\textbf{4.92} \\ \midrule
    \multirow{4}[0]{*}{SQuAD} & No Attack & - & 20.50& 	2.67	& 0.00	& 82.33& 	4.44	& 83.33& 	15.33	& 58.00	& 33.33 \\
          & Dipper & 75.33 & 3.00	&1.33	&1.33	&57.67&	1.00&	50.00&	39.33&	43.33	&24.62 \\
          & Raidar & 95.33 & 21.33&	5.33	&0.00	&59.33	&5.20	&68.00	&8.67	&48.67	&27.07 \\
          & \gc Ours  & \gc 88.67  & \gc \textbf{1.17} & \gc	\textbf{1.33}	 & \gc \textbf{0.00}	 & \gc \textbf{9.67}	 & \gc \textbf{1.00}	 & \gc \textbf{12.67}	 & \gc \textbf{1.33}	 & \gc \textbf{0.00}	 & \gc \textbf{3.40} \\ \midrule
    \multirow{4}[0]{*}{LongQA} & No Attack & -     & 28.50 & 7.50  & 0.00  & 74.00 & 7.25  & 82.67 & 28.67 & 74.33 & 37.87 \\
      & Dipper & 94.67 & 6.50  & \textbf{0.00}  & 0.00  & 55.67 & 1.00  & 31.33 & 49.33 & 63.33 & 25.90 \\
      & Raidar & 100.00 & 15.83 & 2.17  & 0.00  & 48.67 & 6.75  & 64.00 & 15.33 & 70.33 & 27.89 \\
      & \gc Ours  &\gc 95.33 &\gc \textbf{2.67}  & \gc 1.33  &\gc \textbf{0.00}  &\gc \textbf{5.33}  &\gc \textbf{0.00}  &\gc \textbf{8.67}  &\gc \textbf{2.00}  &\gc \textbf{3.00}  &\gc \textbf{2.88} \\ \bottomrule
    \end{tabular}}
  \label{tab:exp_fpr1}%
\end{table*}%


This system prompt is similar to that in \cite{sadasivan2023can} and instructs the LLM to act as a paraphraser while maintaining the text quality of the sentences before the paraphrasing.

\begin{tcolorbox}[title=User Prompt]
Rewrite the following INPUT in the tone of a text message to a friend without any greetings or emojis:
\end{tcolorbox}

Rather than making a direct request for human-toned texts, we experimentally find that presenting a realistic conversation scene with humans can better guide the LLM to produce more vivid and diverse sentences, further boosting stronger attacks. 

(2) For machine-style distributions:

To elicit machine-like responses from LLMs, our prompt engineering (PE) constructs a wide range of prompts, \eg ask for direct paraphrasing or instruct the LLM to reply in the tone of an AI assistant. The LLM likelihood provided by a detector is used as a proxy to quantify the degree of machine-style characteristics in the output. We find that naively prompting an LLM to rewrite a machine text often reduces existing machine-related features, as the rewritten texts inherit mixed stylistic and lexical preferences from multiple LLMs, increasing textual diversity. To preserve machine-related features, we incorporate stylistic cues into the prompts, e.g., instructions like "... in the tone of an AI assistant ..." or "... in a machine-writting style ..." to guide the LLM for a more typical machine expression, which indeed makes the output more easily detectable and, in turn, enhances the attack. 
\begin{tcolorbox}[title=System Prompt]
You are a helpful assistant.
\end{tcolorbox}
\begin{tcolorbox}[title=User Prompt:]
Repeat the following paragraph:
\end{tcolorbox}

Considering that the original LLM-generated sentence carries the richest machine-related characteristics that are most easily detected, we adopt a direct and effective strategy by making LLM repeat the input sentence to obtain the most machine-like token probabilities. We provide the results of representative prompts in Fig. \ref{fig:reduce_by_amplify}, which validate the effectiveness of our selection strategy.

\section{Additional Results}
\label{sec:add_results}

\begin{figure}[t]
\begin{center}
\includegraphics[width=\linewidth]{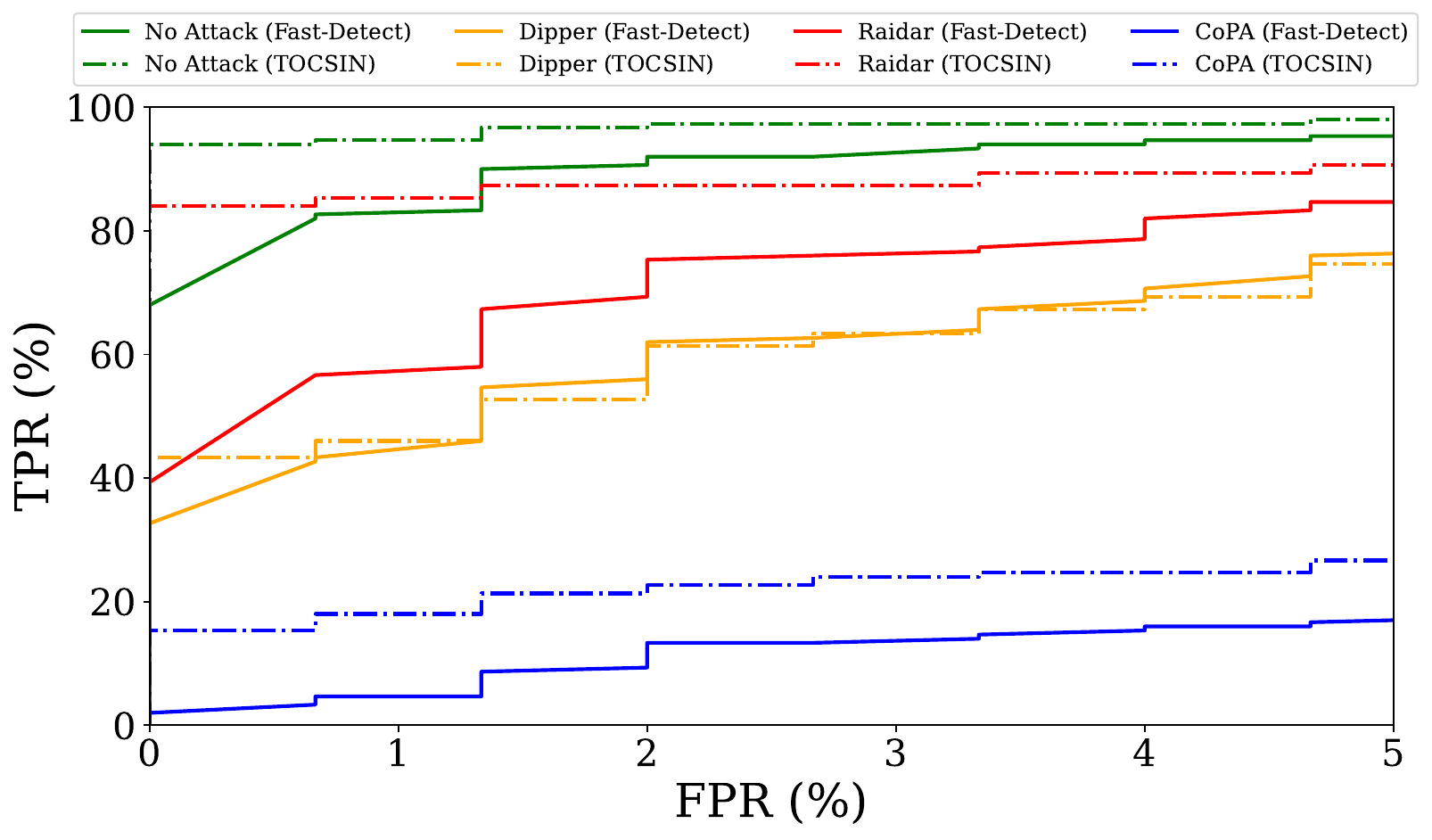}
\end{center}
\vspace{-1em}
\caption{ROC (0-5\% FPR) for GPT-3.5-turbo on Fast-DetectGPT and TOCSIN before and after paraphrasing. The proposed CoPA achieves the best detection rate across various FPRs.}
\label{fig:roc}
\end{figure}

\textbf{More Source LLMs.}
\label{exp:more_source}
We provide the performance of paraphrasing on machine texts generated by GPT-4o \cite{achiam2023gpt} and Gemini-1.5 Pro \cite{team2024gemini} in Table \ref{tab:exp_more_more_source}. It can be observed that the proposed CoPA continues to achieve better attack effectiveness than existing paraphrasing methods. Also, the detection algorithms obtain relatively worse clean performance on texts generated by two advanced LLMs. 
\begin{figure*}[t]
\begin{center}
\includegraphics[width=\linewidth]{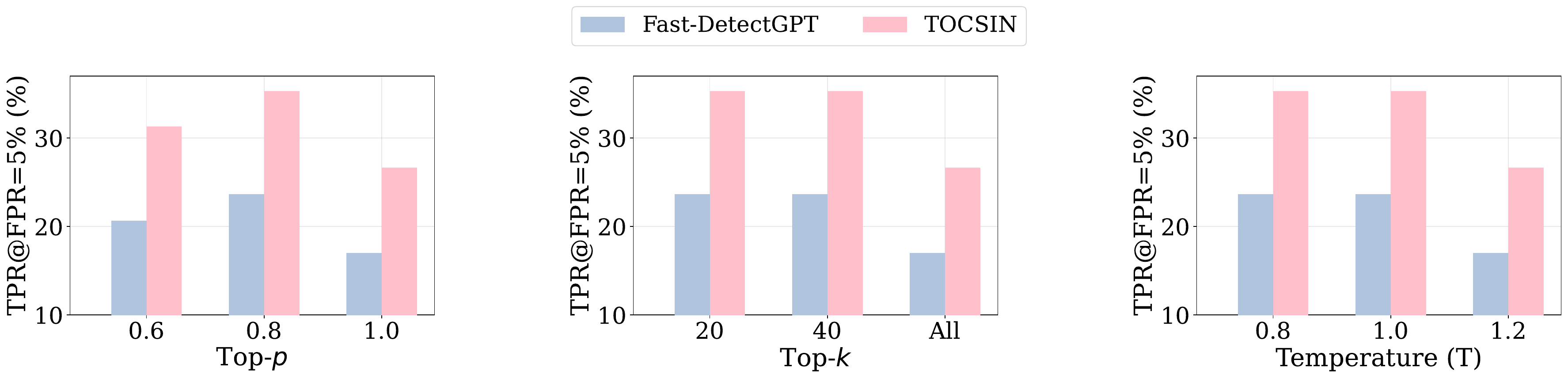}
\end{center}
\vspace{-1em}
\caption{Detection accuracy (at FPR=5\%) of CoPA under different sampling parameters against Fast-DetectGPT \cite{baofast} and TOCSIN \cite{ma2024zero}. We calculate results using samples from the XSum dataset.}
\label{fig:sampling_parameters}
\end{figure*}

\begin{table*}[t]
  \centering
  \caption{Attack results paraphrased by different off-the-shelf LLMs (at 5\% FPR) using GPT-3.5-turbo generated texts from three different datasets.}
    \resizebox{0.88\linewidth}{!}{\begin{tabular}{ccccccc} \toprule
    \multirow{2}[0]{*}{Dataset} & \multirow{2}[0]{*}{Detector} & \multicolumn{5}{c}{Paraphraser} \\ \cmidrule(lr){3-7}
          &       & No Attack & R1-Distill-32B & QwQ-32B & GLM-4-9b-hf & Qwen2.5-72B \\ \midrule
    \multirow{3}[0]{*}{XSum} & Fast-DetectGPT & 95.33 & 24.33 & 11.33  &  31.33 & 17.00 \\
          & TOCSIN & 98.00 & 42.67 & 10.00  &  87.00 & 26.67 \\
          & R-Detect & 49.33 & 6.33  & 00.00   & 38.33 & 0.00 \\ \midrule
    \multirow{3}[0]{*}{SQuAD} & Fast-DetectGPT & 93.50 & 26.83 & 9.50   & 31.33 & 27.50 \\
          & TOCSIN & 92.67 & 42.00 & 17.33  & 81.33 & 25.33 \\
          & R-Detect & 58.00 & 13.00 & 0.00   & 39.67 & 0.00 \\ \midrule
    \multirow{3}[0]{*}{LongQA} & Fast-DetectGPT & 86.00 & 14.67 & 11.33  & 27.33 & 11.33 \\
          & TOCSIN & 88.67 & 26.00 & 2.00  & 78.00 & 16.00 \\
          & R-Detect & 74.33 & 22.67 & 0.00 & 47.33 & 3.00 \\ \bottomrule
    \end{tabular}}
  \label{tab:more_paraphrasers}%
\end{table*}%
\textbf{Results of FPR=1\%.} We then evaluate the attack under a more strict setup where FPR=1\%. As shown in Table \ref{tab:exp_fpr1}, CoPA consistently exhibits superior attack performance over current paraphrasing attacks. We also observe that some detection algorithms fail to produce any defense effects even without any attack at TPR=1$\%$, raising concerns about their feasibility in practical scenarios.

\textbf{ROC Curve Analysis.} Figure \ref{fig:roc} shows the TPR trends corresponding to different FPR values varying from 0\% to 5\%. As observed, TPR generally increases as FPR grows, and CoPA significantly reduces the TPR values of detection algorithms across all FPR thresholds, which strongly validates the effectiveness of the proposed CoPA. Note that under the stricter and more realistic setting of FPR = 1\%, CoPA reaches a detection accuracy below 20\% against these defenses, further underscoring its superior performance.

\textbf{Impact of LLM Sampling Parameters.}
\label{exp:sample_params}
During decoding, LLMs employ various sampling parameters such as Top-$p$, Top-$k$, and the temperature coefficient \(T\) to adjust the sampling results. To investigate their influence, we conduct ablation studies regarding these parameters during the decoding process of our paraphrasing in Figure \ref{fig:sampling_parameters}.  For Top-\(p\) and Top-$k$, the reduction of \(p\) or $k$ results in a drop in sampling diversity since fewer tokens are retained, thereby generating more machine patterns and impairing the performance. For the temperature (\(T\)), numeric results indicate that the increase of \(T\) facilitates the creativity and diversity of sampling choices by reducing the difference in token probabilities, hence better misleading the detectors.  
An adversary can adjust these parameters based on their needs to achieve a superior attack while controlling the writing styles of generated sentences.

\textbf{Results of More LLMs as paraphrasers.}
To validate the universality of the proposed attack, we next consider more LLMs as the paraphraser, including various model scales and recently prevalent reasoning-based models. Specifically, we consider Deepseek R1-Distill-32B \cite{deepseekai2025deepseekr1incentivizingreasoningcapability}, QwQ-32B \cite{qwq32b}, and GLM-4-9B-hf \cite{glm2024chatglm}. The quantitative results in \cref{tab:more_paraphrasers} show the effectiveness of the proposed CoPA across various LLMs. Note that the QwQ-32B generally achieves the best performance. However, the reasoning-based models require significantly more inference time than regular models. We choose the Qwen2.5-72B to balance effectiveness and efficiency.
\begin{table}[htbp]
  \centering
  \caption{Comparison of CoPA with a surrogate-based paraphrasing attack against two SOTA detectors.}
    \resizebox{0.9\linewidth}{!}{\begin{tabular}{cccc} \toprule
    \multirow{2}[0]{*}{Attack} & \multirow{2}[0]{*}{Sim} & \multicolumn{2}{c}{Defense} \\ \cmidrule(lr){3-4}
          &       & Fast-DetectGPT & TOCSIN \\ \midrule
    RedTeaming & 78.00    & 30.00    & 38.67 \\
    Ours  & \textbf{94.00} & \textbf{17.00} & \textbf{26.67} \\ \bottomrule
    \end{tabular}}
  \label{tab:red_teaming}%
\end{table}%

\begin{figure*}[t]
\begin{center}
\includegraphics[width=\linewidth]{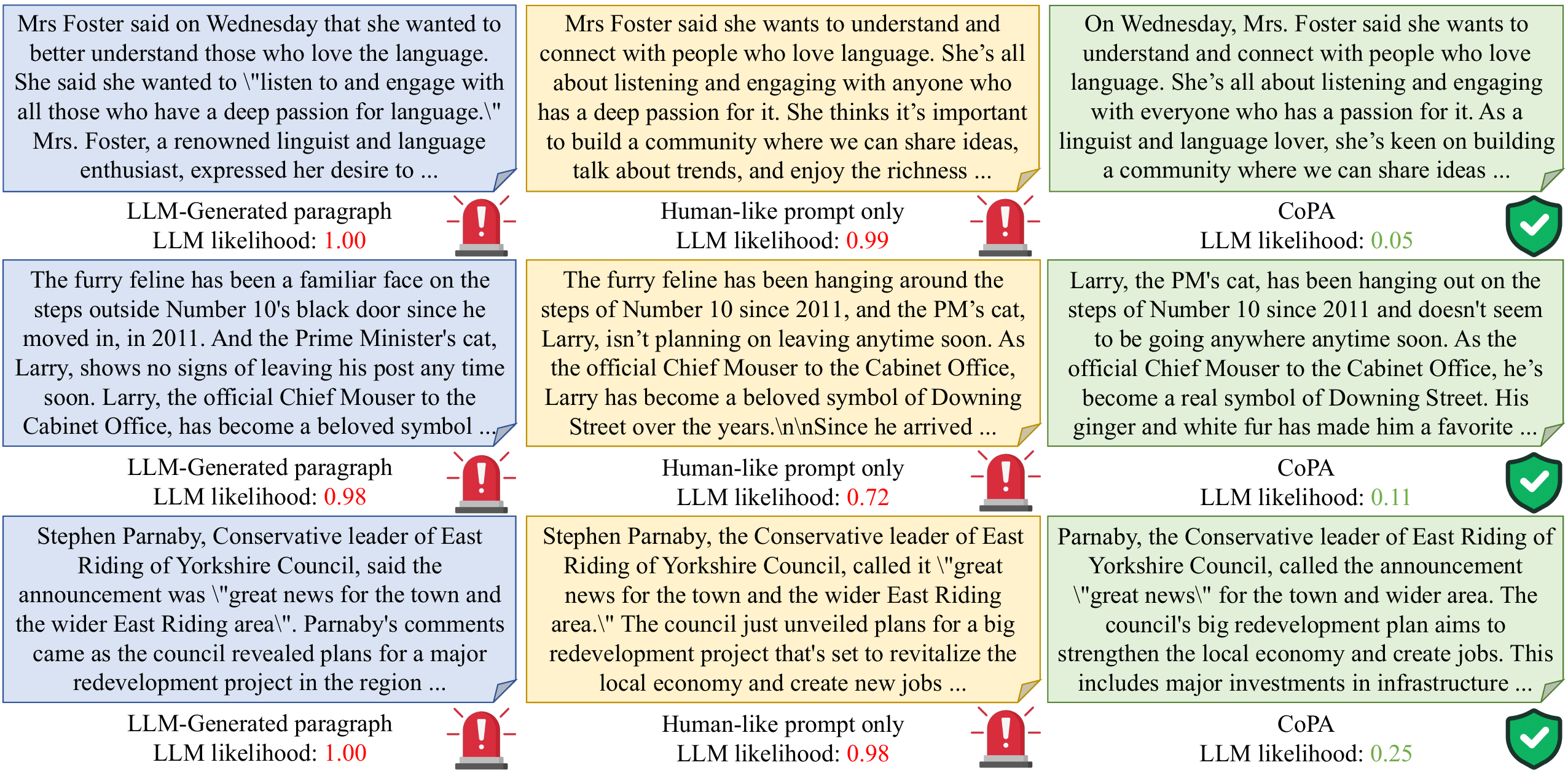}
\end{center}
\vspace{-1em}
\caption{Visualization of paraphrased sentences from human-like distribution $p_h$ and our contrastive distribution $p_c$ (\ie our CoPA). The LLM likelihood is calculated based on Fast-DetectGPT.}
\label{fig:human_only}
\end{figure*}
\textbf{Comparison with a surrogate-based baseline.} This paper follows Dipper \cite{krishna2023paraphrasing} and focuses on the more practical and universal attacks without relying on any surrogate detection model. A direct comparison of these methods with the surrogate-based paraphrasing attack introduced in RedTeaming~\cite{shi2024red} may raise concerns of unfairness.
However, results in Table \ref{tab:red_teaming} reveal that the proposed CoPA can still achieve better performance than RedTeaming.

Notably, we include these results only for experimental completeness. The surrogate-based methods are not the focus of this work.

\textbf{Ablation study of the adaptive truncation mechanism.} To avoid penalizing the probabilities of reasonable and valid tokens, we incorporate an adaptive truncation mechanism to constrain the output token distribution.
We provide an ablation study to investigate its influence.
As observed in Table \ref{tab:ablation_atm}, by truncating the token distribution within a reliable token candidate pool, we can improve the text quality of the generated sentences while maintaining attack effectiveness.

\begin{table}[htbp]
  \centering
  \caption{Ablation on the adaptive truncation technique.}
  
    \resizebox{\linewidth}{!}{\begin{tabular}{ccccc} \toprule
    Method & \multicolumn{1}{l}{Sim} & \multicolumn{1}{l}{Fast-DetectGPT} & \multicolumn{1}{l}{TOCSIN} & \multicolumn{1}{l}{R-Detect} \\ \midrule
    w/o truncation & 93.33  & 19.67  & \textbf{31.33}  & 5.33  \\
    CoPA  & \textbf{94.00 } & \textbf{17.00 }  & 26.67  & \textbf{4.67} \\ \bottomrule
    \end{tabular}}
  \label{tab:ablation_atm}%
\end{table}%



\section{Analysis of human-like prompt only}
\label{examples_human_only}
As shown in Figure \ref{fig:human_only}, we observe that solely relying on the human-like prompt $x_h$ results in unstable attack performance, \ie some sentences derived from the human-like distribution $p_h$ still retain prominent machine-related features, which render them easily identifiable by text detectors. To alleviate this issue, our CoPA framework utilizes an auxiliary machine-like distribution to fully remove these machine characteristics from $p_h$, significantly deceiving text detectors and leading to incorrect predictions. As corroborated by more detailed empirical studies, the proposed contrastive strategy greatly boosts the effectiveness and stability of the paraphrasing attack.

\section{Details about Machine Prompt}
\label{sec:prompt_analysis}
The prompts used in \cref{fig:reduce_by_amplify} are as follows:
\begin{tcolorbox}[title=Machine Prompt 1]
Repeat the following paragraph:
\end{tcolorbox}
\begin{tcolorbox}[title=Machine Prompt 2]
Rewrite the following paragraph in the tone of an AI assistant: 
\end{tcolorbox}
\begin{tcolorbox}[title=Machine Prompt 3]
Paraphrase the following paragraph: 
\end{tcolorbox}
\begin{tcolorbox}[title=Machine Prompt 4]
Rewrite the following paragraph:
\end{tcolorbox}

\section{An Adaptive Defense Strategy}
\label{sec:adpative_defense}
To alleviate the proposed threat, we implement an adaptive defense that adversarially trains OpenAI's LLM-text classifier RoBERTa-large. We fine-tune the model for 10 epochs using 5k human texts and 5k machine texts (including both the original machine texts and those paraphrased by CoPA). We present the optimal performance at a proportion of 50\% CoPA-paraphrased samples within the 5k machine texts in \cref{tab:adpative_defense}.

\begin{table}[!htbp]
  \centering
  \caption{Detection accuracy (at 5\% FPR) of texts generated by GPT-3.5-turbo based on the XSum dataset.}
    \resizebox{0.86\linewidth}{!}{\begin{tabular}{ccc} \toprule
    Attack & w/o training & Adversarial training \\
    \midrule
    No Attack & 66.67 & 99.78 \\

    CoPA  & 22.67 & 78.00 \\ \bottomrule
    \end{tabular}}
  \label{tab:adpative_defense}%
\end{table}%

The results indicate that the adaptive defense based on texts provided by our CoPA can alleviate the proposed threat to some extent.

\section{Human Prompts for Different Contexts}
To better handle the different demands of contexts, we conduct a preliminary study by designing two alternative prompts as follows:

\begin{tcolorbox}[title=Human Prompt I for General Scenarios]
Rewrite the following INPUT in human style, with varying sentence structures and replace common terms with nuanced synonyms. Maintain the meaning and avoid any repetition.
\end{tcolorbox}

\begin{tcolorbox}[title=Human Prompt II for Academic Context]
Rewrite the following INPUT in a human-written academic style, with varying sentence structures and replace common terms with nuanced synonyms. Maintain the meaning and avoid any repetition. 
\end{tcolorbox}

Here are the corresponding attack results:

\begin{table}[htbp]
  \centering
  \caption{Attack results (at 5$\%$ FPR) of different human prompts using GPT-3.5-turbo generated texts on XSum.}
    \resizebox{0.98\linewidth}{!}{\begin{tabular}{ccccc} \toprule
    Prompt & Fast-DetectGPT & TOCSIN & RoBERTa & R-Detect \\ \midrule
    Prompt I & 24.67 & 43.33  & 15.33 & 18.67 \\
    Prompt II & 24.33 & 39.67 & \textbf{5.33}  & 14.33 \\
    CoPA  & \textbf{17.00} & \textbf{26.67} & 22.67 & \textbf{4.67} \\ \bottomrule
    \end{tabular}}
  \label{tab:addlabel}%
\end{table}%

The quantitative results in Tab. 1 and demonstrations of paraphrased sentences show that CoPA is able to flexibly combine with prompts of various styles while maintaining high effectiveness. Users may design their own prompts based on our provided template to adapt to diverse writing styles. 

\section{Evaluation on Text Quality}
\label{sec:text_quality}
Apart from the attack effect, it is necessary to analyze the text quality of paraphrased sentences. Specifically, we first conduct a deeper linguistic analysis of the output texts in Table \ref{tab:linguistic}, including TTR (Type-Token Ratio) and MTLD (Measure of Textual Lexical Diversity) metrics for lexical diversity, Div$\_$syn \cite{guo2024curious} for syntactic variability. The results demonstrate the high quality of our generated sentences.

\begin{table}[htbp]
  \centering
  \caption{Analysis of paraphrased sentences from different methods based on the XSum dataset.}
    \resizebox{0.78\linewidth}{!}{\begin{tabular}{cccc} \toprule
    Method & TTR↑  & MTLD↑ & Div$\_$syn↑  \\ \midrule
    No Attack & 0.5734 & 111.4913 & 0.4700 \\
    Dipper & 0.5511 & 74.8257 & 0.4648 \\
    CoPA  & 0.6746 & 145.8134 & 0.5593 \\ \bottomrule
    \end{tabular}}
  \label{tab:linguistic}%
\end{table}%

\begin{table*}[!t]
  \centering
  \caption{Comparison of our method with Dipper on text quality. The perplexity is calculated on GPT-neo.}
    \resizebox{0.97\linewidth}{!}{\begin{tabular}{cccccccc} \toprule
    \multirow{2}[0]{*}{Dataset} & \multirow{2}[0]{*}{Text} & \multirow{2}[0]{*}{Sim$\uparrow$} & \multirow{2}[0]{*}{Perplexity$\downarrow$} & \multicolumn{2}{c}{GPT-4 Eval} & \multicolumn{2}{c}{Human Eval} \\  \cmidrule(lr){5-8}
          &       &       &       & Natural fluency$\uparrow$ & Consistency$\uparrow$ & Natural fluency$\uparrow$ & Consistency$\uparrow$ \\ \midrule
    \multirow{4}[0]{*}{XSum} & Human & -     & 16.11 & 3.88  & -     & 4.40   & - \\ 
          & Machine & -     & 8.857 & 4.33  & -     & 4.94  & - \\
          & Dipper & 86.67 & 14.76 & 3.74  & 3.76  & 4.25  & 4.26 \\
          & Ours  & 94    & 15.58 & 4.64  & 4.95  & 4.74  & 4.87 \\ \midrule
    \multirow{4}[0]{*}{SQuAD} & Human & -     & 19.52 & 3.60   & -     & 3.98  & - \\
          & Machine & -     & 10.28 & 4.71  & -     & 4.81  & - \\
          & Dipper & 75.33 & 14.70 & 3.53  & 3.57  & 4.02  & 4.05 \\
          & Ours  & 88.67 & 17.77 & 4.56  & 4.91  & 4.56  & 4.79 \\ \midrule
    \multirow{4}[0]{*}{LongQA} & Human & -     & 27.54 & 3.48  & -     & 3.75  & - \\ 
          & Machine & -     & 7.79 & 4.91  & -     & 4.99  & - \\
          & Dipper & 94.67 & 11.61 & 3.57  & 4.11  & 4.04  & 4.23 \\
          & Ours  & 95.33 & 13.08 & 4.57  & 4.99  & 4.53  & 4.81 \\ \bottomrule
    \end{tabular}} 
  \label{tab:text_quality}%
\end{table*}%

We also provide a comprehensive evaluation of natural fluency and semantic consistency with additional key metrics, including text perplexity, GPT4-assisted evaluation, and human study. To conduct the human evaluation, we choose GPT-3.5-turbo as the source model and randomly select 100 pairs of texts from each dataset for human annotators. The evaluation criteria generally align with those in Dipper \cite{krishna2023paraphrasing}, where we recruit 10 native English speakers from Amazon Mechanical Turk (MTurk) to perform the evaluation. We report the average scores to reduce subjective biases in \cref{tab:text_quality}. The results indicate that CoPA produces paraphrased texts with lower perplexity than authentic human-written texts, while achieving substantially better fluency and semantic consistency compared to those generated by Dipper, in both terms of GPT-4 assited and human evaluation.

The instructions given to human annotators for \textit{semantic consistency} align with those in Dipper, while we provide the scoring standard for \textit{natural fluency} and the detailed prompt for GPT-4 auto evaluation as follows:
\newpage
\begin{tcolorbox}[title=Instruction for Human evaluation:]
Natural Fluency Scoring (1–5)
\\

5. Excellent: Text flows perfectly naturally with varied, idiomatic phrasing. Grammar, word choice, and sentence structure appear completely native with zero awkwardness.
\\

4. Good: Text reads smoothly with only minor and infrequent awkwardness. May contain 1-2 subtle non-native phrasings, but remains highly readable.
\\

3. Fair: Generally understandable but contains noticeable unnatural phrasing. Some grammatical errors or awkward constructions occasionally disrupt flow.
\\

2. Poor: Frequent unnatural phrasing and grammatical errors make reading difficult. Requires effort to understand in places.
\\

1. Very Poor: Severely broken or unnatural English with major grammar issues. Often difficult or impossible to understand.
\end{tcolorbox}

\begin{tcolorbox}[float*=b, width=\textwidth, title=Prompt for GPT-4 evaluation:]
(1) Task: Evaluate the natural fluency of a given sentence. Use a 5-point scale (5 = highest).

Natural Fluency (1-5):

\hspace*{2em}1. Does the rewritten sentence flow naturally, avoiding awkward phrasing or redundancy?

\hspace*{2em}2. Assess grammar, word choice, and readability (e.g., smooth transitions between clauses).

\hspace*{2em}3. Penalize unnatural idioms or register mismatches (e.g., mixing formal and colloquial terms)
\\

Output Format

Please provide the score for the metric.

Include a concise rationale (1-2 sentences per metric) highlighting specific strengths/weaknesses.

Example:

      \hspace*{2em} INPUT: "The deadline got pushed back because of unexpected tech issues."
      
      \hspace*{2em} OUTPUT: 4/5 (Colloquial tone matches intent; "pushed back" is natural but "tech issues" slightly informal).
\\

(2) Task: Evaluate the semantic consistency of a rewritten sentence compared to its original version. Use a 5-point scale (5 = highest).

Semantic Consistency (1–5):

\hspace*{2em}1. Does the rewritten sentence preserve the original meaning?

\hspace*{2em}2. Check for critical information retention, logical coherence, and absence of distortion.

\hspace*{2em}3. Deduct points for omissions, additions, or ambiguous interpretations
\\

Output Format

Provide the score for the metric.

Include a concise rationale (1–2 sentences per metric) highlighting specific strengths/weaknesses.

Example:

      \hspace*{2em}INPUT:
      
            \hspace*{4em}Original: "The project deadline was extended due to unforeseen technical challenges."
            
            \hspace*{4em}Rewritten: "The deadline got pushed back because of unexpected tech issues."
            
      \hspace*{2em}OUTPUT:
      
            \hspace*{4em}5/5 (Key details retained; no loss of meaning).
\end{tcolorbox}

\end{document}